\documentclass{article} % For LaTeX2e
\usepackage{iclr2026_re-align_workshop,times}

\usepackage{amsmath,amsfonts,bm}

\def\eqref#1{equation~\ref{#1}}
\def\1{\bm{1}}

\DeclareMathAlphabet{\mathsfit}{\encodingdefault}{\sfdefault}{m}{sl}
\SetMathAlphabet{\mathsfit}{bold}{\encodingdefault}{\sfdefault}{bx}{n}

\usepackage{hyperref}
\usepackage{url}

\usepackage{float}
\usepackage{graphicx} % For including images
\usepackage{subfigure}  % light-weight; load after graphicx
\usepackage{booktabs}
\usepackage{makecell}
\usepackage{siunitx}
\usepackage{enumitem}

\usepackage{algorithm} 
\usepackage{algorithmic}

\usepackage{amsfonts}        % blackboard bold
\usepackage{mathtools}       % fixes to amsmath
\usepackage{nicefrac}        % compact 1/2 etc.

\usepackage[dvipsnames]{xcolor} % for BrickRed, ForestGreen, etc.
\usepackage{pifont}             % for \ding symbols
\newcommand{\cmark}{\textcolor{ForestGreen}{\ding{51}}} % ✓
\usepackage{tikz}
\usepackage{xcolor}
\usepackage{CJKutf8}

\newcommand{\zh}[1]{\begin{CJK}{UTF8}{gbsn}#1\end{CJK}}
\definecolor{encolor}{HTML}{4A90D9}    % blue for English
\definecolor{zhcolor}{HTML}{D94A4A}    % red for Chinese
\definecolor{escolor}{HTML}{D9944A}    % orange for Spanish

\usepackage{xcolor}
\usepackage{soul}

\definecolor{encolor}{HTML}{D6E4F0}
\definecolor{enborder}{HTML}{4A90D9}
\definecolor{zhcolor}{HTML}{F0D6D6}
\definecolor{zhborder}{HTML}{D94A4A}
\definecolor{escolor}{HTML}{F0E2D6}
\definecolor{esborder}{HTML}{D9944A}

\usepackage{xcolor}
\usepackage{tcolorbox}
\usepackage{amssymb} % For the small triangle

\definecolor{hl-exhibit}{HTML}{FFF4C1} % Yellow-ish
\definecolor{hl-announce}{HTML}{E0F2F1} % Teal/Green-ish
\definecolor{hl-finance}{HTML}{E8EAF6} % Blue-ish
\definecolor{hl-nutrition}{HTML}{FFECB3} % Orange-ish
\definecolor{hl-transport}{HTML}{FFEBEE} % Red-ish
\definecolor{hl-brand}{HTML}{F3E5F5}    % Violet-ish

\usepackage[capitalize,noabbrev]{cleveref}  % load AFTER hyperref (already in class)

\title{Polysemantic Experts, Monosemantic Paths: Routing as Control in MoEs}

\author{Charles Ye\thanks{Corresponding author.}\\
Independent\\
\texttt{dogdynamics@proton.me}
\texttt{} \\
\And
Bo Yuan \\
Georgia Institute of Technology \\
\texttt{byuan48@gatech.edu} \\
\And
Lee Sharkey \\
Goodfire \\
\texttt{lee@goodfire.ai}
}

\iclrfinalcopy % Uncomment for camera-ready version, but NOT for submission.
\begin{document}

\maketitle

\begin{abstract}
An LLM's residual stream is both \textbf{state} and \textbf{instruction}: it encodes the current context and determines the next transformation. We introduce a parameter-free decomposition for Mixture-of-Experts models that splits each layer's hidden state into a \textbf{control} signal that causally drives routing and an orthogonal \textbf{content} channel invisible to the router. Across six MoE architectures, we find that models preserve surface-level features (language, token identity, position) in the content channel, while the control signal encodes an abstract function that rotates from layer to layer. Because each routing decision is low-bandwidth, this rotation forces compositional specialization across layers. While individual experts remain polysemantic, \textbf{expert paths become monosemantic}, clustering tokens by semantic function across languages and surface forms. The same token (e.g., ``:'') follows distinct trajectories depending on whether it serves as a type annotation, an introductory colon, or a time separator. Our decomposition identifies the source of this structure: clusters in the control subspace are substantially more monosemantic than those in the full representation. As a result, the natural unit of interpretability in MoEs is not the expert but the trajectory.
\end{abstract}

\section{Introduction}

\label{sec:intro}
Mixture-of-Experts (MoE) architectures have become the  dominant paradigm for frontier language  models\footnote{At time of writing, all top-10 models with disclosed architectures on \textsc{ArtificialAnalysis}~\citep{ArtificialAnalysis2025}  are MoEs.}. While the mechanism is architecturally simple, how MoEs organize computation remains poorly understood.

The intuitive model of a ``committee of specialists'' where each expert masters a distinct domain has been  largely invalidated \citep{jiang2024mixtral, xue2024openmoe}. Instead, experts are  \emph{polysemantic}, activating for unrelated concepts. Prior work finds varied and often weak correlations 
between experts and interpretable categories: semantic domains \citep{olson2025probingsemanticroutinglarge}, POS \citep{antoine2024partofspeechsensitivityroutersmixture}, or sequence position \citep{bershatsky2025spatialstructuremixtureofexpertstransformers}. Such  disparate findings suggest that individual experts may simply be the wrong unit of analysis.

\begin{figure}[t]
    \centering
    \includegraphics[width=.95\linewidth]{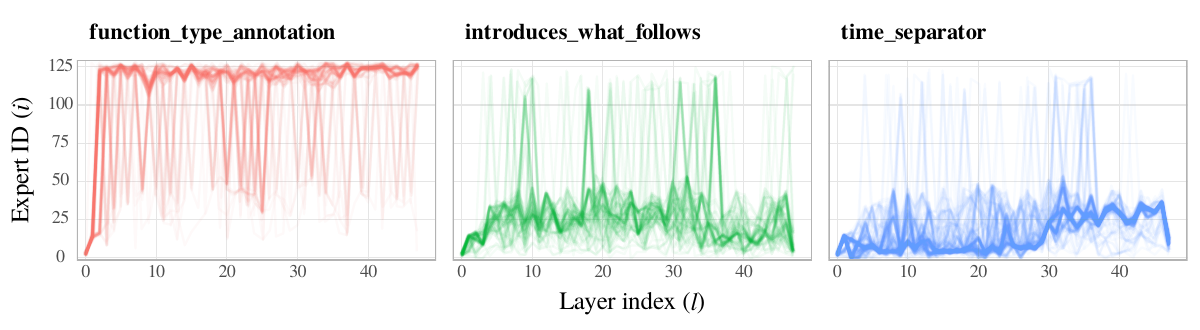}
    \vspace{-0.5em}
    \caption{\textbf{One token, three programs.} We plot 500 top-1 routing paths of the token \smash{\fcolorbox{black}{cyan!25}{\small:}} through 48 MoE layers of \textsc{Qwen3-30B-A3B}, separated by contextual function: Python type annotation (\emph{left}), introductory colon (\emph{center}), time separator (\emph{right}). Despite identical token ID, the three uses produce distinct expert trajectories. 
    %Early layers share similar routing; divergence emerges as the model resolves contextual meaning across depth.
    Expert IDs are reordered at each layer using a shared Sugiyama-style layout to minimize edge crossings (\Cref{app:sugiyama}). The same expert ordering is used for all panels.}
    \label{fig:colon}
\end{figure}

Tracking which expert a token visits at a single layer yields little structure. But we find that tracking the \emph{sequence} of experts across layers tells a different story. \Cref{fig:colon} shows the routing paths of 500 instances of the token \smash{\fcolorbox{black}{cyan!25}{\small:}} in three distinct syntactic roles\footnote{Sequences are generated for balanced category representation; samples in \Cref{app:colon-generation}.}. Despite identical surface form, the three uses follow visibly distinct trajectories: coherent bundles that diverge based on contextual function, not token identity. Individual experts are polysemantic; expert \textbf{paths} are not.

What mechanism produces this structure, and why has prior work missed it? Existing analyses primarily study correlations between routing and external variables, without isolating what \emph{causally drives} expert selection. Routing is fully determined by $R_l^{\top}h_l$, the product of a routing matrix and the hidden state $h_l$. Since $h_l$ encodes language, syntax, and semantics \citep{bert2019}, correlations between routing and such variables are unsurprising. But this linearity is not just a limitation---it is an opportunity.

An LLM's residual stream is both \textbf{state} and \textbf{instruction}: it encodes the context being processed, and it determines what computation is applied next. In dense models these roles are entangled. MoEs make them separable: because routing is a linear read of $h_l$, we can derive exactly which components causally influence expert selection (the \textbf{control} signal) and which are invisible to the router (the \textbf{content} channel). This decomposition lets us trace how MoEs organize computation, from the geometry of routing to the monosemantic paths of \Cref{fig:colon}.

Our contributions:
\begin{itemize}[leftmargin=1.5em]
    \item \textbf{A causal decomposition of the residual stream} (\Cref{sec:svd-decomp}). We derive a parameter-free decomposition that separates each layer's hidden state into orthogonal \emph{control} (router-visible) and \emph{content} (router-blind) channels. This enables direct analysis of what information causally drives expert selection versus what is merely carried through.
    \item \textbf{A rotating computational strategy} (\Cref{sec:computation}).  Across 6 models, we find a shared strategy: routing is dominated by a tiny subspace of the hidden state that rotates from layer to layer. This rotation mechanism  forces compositional specialization.
    \item \textbf{Expert paths are monosemantic} (\Cref{sec:paths}). While individual experts remain polysemantic, multi-layer expert paths cluster 
    tokens by semantic function across languages and surface forms. Control-subspace clusters group ${4}{\times}$ more unique tokens than content 
    clusters at equal interpretability, confirming that the decomposition isolates the signal responsible for this structure.
\end{itemize}

\section{Separating State from Instruction}
\label{sec:svd-decomp}
Which features of the residual stream actually drive routing? Routing is a linear function of the residual stream: $s_l = R_l^{\top}h_{l}$. This linearity is crucial—it means the router can only access information that lies in the row space of its weight matrix $R_l$. Anything orthogonal to that subspace is invisible to routing, regardless of what it encodes. We exploit this to exactly separate each layer's residual stream $h_l$ into two components: what the router can read, and what it cannot.

\begin{align*}
  h_{l}
  \;=\;
  \underbrace{h^{\text{vis}}_{l}}_{\text{router-visible}}
  \;+\;
  \underbrace{h^{\text{blind}}_{l}}_{\text{router-blind}},
  \qquad
  \langle h^{\text{vis}}_{l},\,h^{\text{blind}}_{l}\rangle = 0.
\end{align*}

The decomposition is straightforward: we compute the SVD of the routing matrix $R_l$ and project $h_l$ onto its row space to obtain $h^{\text{vis}}_l$; the remainder is $h^{\text{blind}}_l$ (full details in \Cref{app:svd-decomp}).

\begin{figure}[t]
    \centering
    \includegraphics[width=0.95\linewidth]{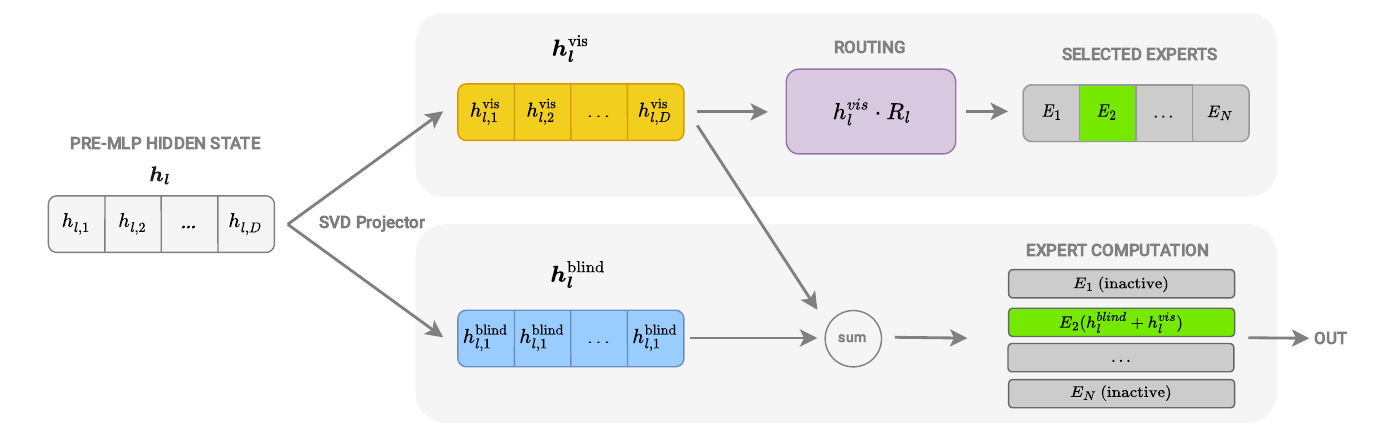}
    \vspace{-0.5em}
    \caption{\textbf{Control and content in an MoE layer.} We decompose the residual stream $h_l$ into orthogonal components via SVD of the routing matrix. The router-visible component ($h^{\text{vis}}_l$, top) is the \textit{control} signal: it alone determines expert selection. The router-blind component ($h^{\text{blind}}_l$, bottom) is the \textit{content}: invisible to routing, but processed by the selected expert alongside 
    $h^{\text{vis}}_l$.}
\label{fig:moe-diagram} 
\end{figure}

By construction, $R_l^T h^{\text{blind}}_l = 0$: the router-blind component \textbf{cannot} influence expert selection. This is not a statistical claim, but a mathematical guarantee. \Cref{fig:moe-diagram} illustrates how the two components flow through the MoE layer.

We validate the decomposition empirically in \Cref{sec:computation}, where it reveals a surprising structure: a control signal that rotates across layers 
while content accumulates.

\section{How MoEs Organize Computation}
\label{sec:computation}
We apply our decomposition across six modern MoE architectures spanning 7B--106B total parameters, with 32--128 experts per layer and diverse designs including shared experts, hybrid attention, and varied sparsities. All analyses use $\sim$2m tokens sampled from \textsc{C4} and \textsc{HPLT} (\Cref{app:models} details models and data). Across all models, we find the same computational motif.

\paragraph{Routers amplify, they don't integrate.} How does a router distill a high-dimensional residual stream $h_l$ into expert choices? One might expect it to integrate subtle signals distributed across the full representation.

Instead, routers converge on a simpler strategy: they amplify what is already loud. Across all models and layers, router weights concentrate on the dimensions of $h_l$ that already carry the highest activation magnitude ($\rho \approx 0.6$; full analysis in \Cref{app:low-rank-routing}). This compresses the effective decision well below the router's capacity: probes trained on just the top \textbf{2\%} of hidden dimensions (ranked by activation magnitude) predict the 
top-1 expert with \textbf{38--78\%} accuracy; random subsets of the same size achieve \textbf{1--4\%}. This compression has consequences: no single routing decision can express fine-grained specialization.

\begin{figure}[!t]
    \centering
    \includegraphics[width=\linewidth]{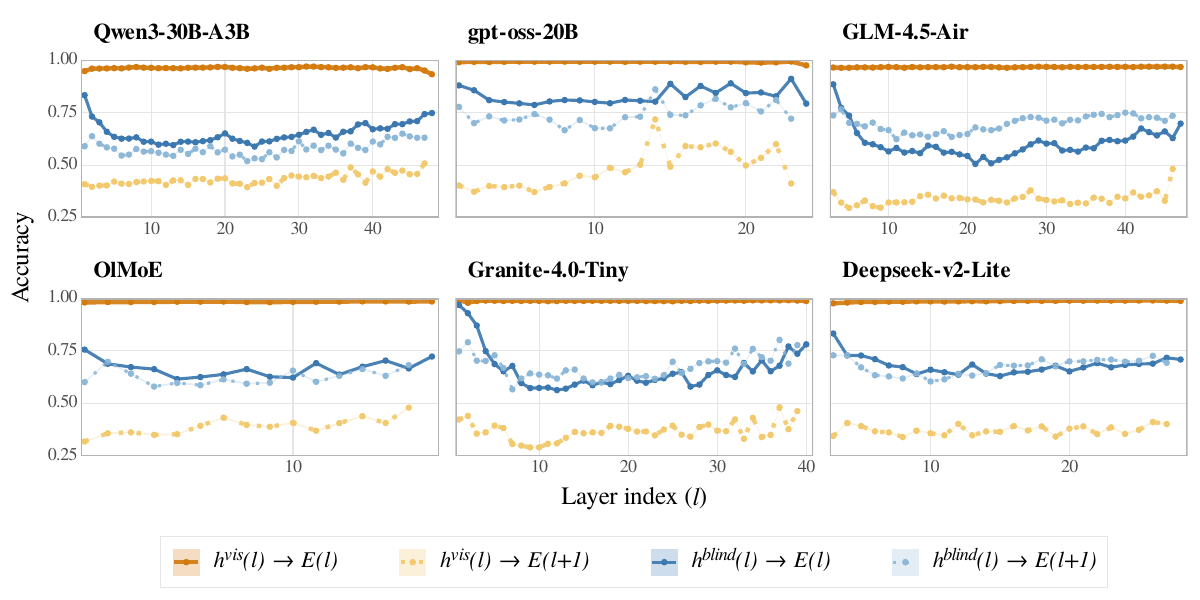}
    \vspace{-1em}
    \caption{\textbf{Routing features are ephemeral.} Probes predict the top-1 expert at the current layer $E(l)$ or next layer $E(l+1)$ from each channel. The control signal $h^{\text{vis}}_l$ predicts $E(l)$ near-perfectly (dark orange, $\sim$99\%) but carries almost no information about $E(l+1)$ (light orange, $\sim$35\%). The content channel $h^{\text{blind}}_l$ shows the reverse: weak for the current layer (dark blue, $\sim$67\%), it is the strongest predictor of the next layer's choice (light blue, $\sim$65\%).}    
    \label{fig:expert-id-predictivity}
\end{figure}

\paragraph{Control rotates; content accumulates.} If routing is driven by a thin slice of $h_l$, what role does the rest play? Our decomposition lets us answer 
this directly, separating features that causally drive routing from those that cannot influence it\footnote{We probe the top-1 expert throughout.}.

We train probes to predict expert choice at the current layer $E(l)$ and the next layer $E(l+1)$ from each channel (\Cref{fig:expert-id-predictivity}). The 
results reveal a clear information pipeline. $h^{\text{vis}}_l$ predicts $E(l)$ near-perfectly ($\sim$99\%), validating the decomposition: the control signal is causally sufficient. But it contains almost no information about $E(l+1)$ ($\sim$35\%)---what drives routing at one layer becomes irrelevant at the next.

The pattern flips for content. $h^{\text{blind}}_l$, which \emph{cannot} influence routing at layer $l$, is the strongest predictor of the next layer's expert 
choice ($\sim$65--75\%). This is the hand-off: \textbf{experts transform content to prepare the next control signal.} Control is continually re-derived from 
content. Cross-layer cosine similarity confirms the asymmetry: $h^{\text{vis}}_l$ changes substantially between adjacent layers while $h^{\text{blind}}_l$ remains stable across depth (\Cref{app:feature_rotation}).

\paragraph{Routing bypasses surface features.} What does the control signal actually encode? We probe both channels for language, token ID, and sequence 
position, features commonly studied in prior work on expert specialization.

Across all six models and every layer, these features reside primarily in $h^{\text{blind}}_l$, not $h^{\text{vis}}_l$ (\Cref{app:channel_probes}). The disparity is largest at middle layers, where surface features are nearly absent from the control signal; they re-enter only at the earliest and final layers, where routing interfaces with token-level representations. Language, token identity, and position are carried in content but do not drive expert selection through the core of the network\footnote{This may explain findings in \citet{moee2025}, where routing-derived embeddings boost $h_l$-derived embeddings: routing strips away surface features, revealing core semantic data.}.

This resolves a puzzle from prior work: routing correlates only weakly with such features \emph{because it is not reading them}. Individual routing decisions are coarse---each one collapses many tokens onto the same expert, discarding the detail that makes $h_l$ rich. But because control rotates, each layer abstracts along a different axis. As we show next, these coarse decisions compose across 
layers into precise semantic signatures.

\section{Expert Paths as Monosemantic Representations}
\label{sec:paths}

If each routing decision is a coarse categorization, and each layer categorizes along a different axis, then sequences of routing decisions should compose into 
fine-grained semantic signatures. We test this by examining \textbf{expert paths}: for a contiguous band of $L$ layers, each token's path is the tuple of $L$ top-1 expert IDs it traverses. We group tokens sharing identical paths and examine the resulting clusters.

\paragraph{Paths cluster by meaning, not surface form.}  We collect paths over ${\sim}$10M multilingual tokens (EN/ZH/ES) from \textsc{C4} and \textsc{HPLT}, using \textsc{gpt-oss-20b} across mid-layers $l = 8, \dots, 16$---the depth range where surface features are most absent from the control signal (\Cref{sec:computation}).

\begin{figure*}[t]
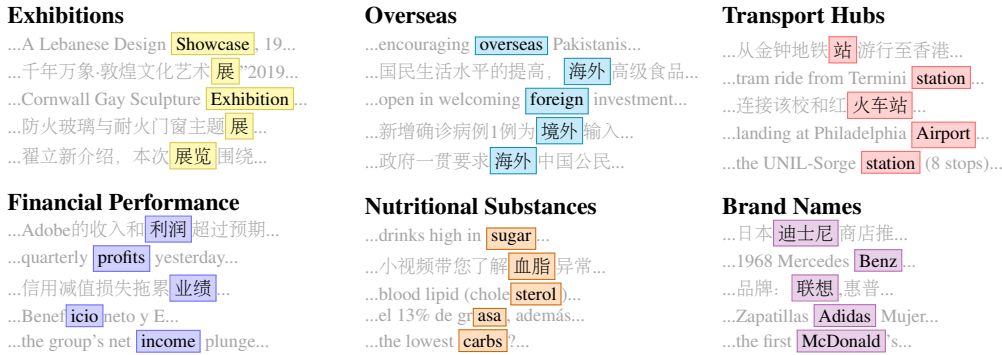

\centering
\scriptsize
\setlength{\fboxsep}{1.5pt}
\newcommand{\ctx}[1]{\textcolor{gray!70}{#1}}

% Softer borders matching fill colors
\newcommand{\hlyellow}[1]{\fcolorbox{yellow!80!black}{yellow!35}{#1}}
\newcommand{\hlblue}[1]{\fcolorbox{blue!60}{blue!18}{#1}}
\newcommand{\hlcyan}[1]{\fcolorbox{cyan!70!black}{cyan!20}{#1}}
\newcommand{\hlorange}[1]{\fcolorbox{orange!80!black}{orange!25}{#1}}
\newcommand{\hlred}[1]{\fcolorbox{red!60}{red!18}{#1}}
\newcommand{\hlviolet}[1]{\fcolorbox{violet!60}{violet!18}{#1}}

\begin{minipage}[t]{0.33\textwidth}
\raggedright

{\footnotesize\textbf{Exhibitions}} \\[2pt]
\ctx{...A Lebanese Design} \hlyellow{Showcase}\ctx{, 19...} \\
\ctx{...\zh{千年万象·敦煌文化艺术}}\hlyellow{\zh{展}}\ctx{\zh{"2019}...} \\
\ctx{...Cornwall Gay Sculpture} \hlyellow{Exhibition}\ctx{...} \\
\ctx{...\zh{防火玻璃与耐火门窗主题}}\hlyellow{\zh{展}}\ctx{...} \\
\ctx{...\zh{翟立新介绍，本次}}\hlyellow{\zh{展览}}\ctx{\zh{围绕}...}

\vspace{8pt}
{\footnotesize\textbf{Financial Performance}} \\[2pt]
\ctx{...\zh{Adobe的收入和}}\hlblue{\zh{利润}}\ctx{\zh{超过预期}...} \\
\ctx{...quarterly} \hlblue{profits} \ctx{yesterday...} \\
\ctx{...\zh{信用减值损失拖累}}\hlblue{\zh{业绩}}\ctx{...} \\
\ctx{...Benef}\hlblue{icio}\ctx{ neto y E...} \\
\ctx{...the group's net} \hlblue{income} \ctx{plunge...}

\end{minipage}%
\hfill%
\begin{minipage}[t]{0.33\textwidth}
\raggedright

{\footnotesize\textbf{Overseas}} \\[2pt]
\ctx{...encouraging} \hlcyan{overseas} \ctx{Pakistanis...} \\
\ctx{...\zh{国民生活水平的提高，}}\hlcyan{\zh{海外}}\ctx{\zh{高级食品}...} \\
\ctx{...open in welcoming} \hlcyan{foreign} \ctx{investment...} \\
\ctx{...\zh{新增确诊病例1例为}}\hlcyan{\zh{境外}}\ctx{\zh{输入}...} \\
\ctx{...\zh{政府一贯要求}}\hlcyan{\zh{海外}}\ctx{\zh{中国公民}...}

\vspace{8pt}
{\footnotesize\textbf{Nutritional Substances}} \\[2pt]
\ctx{...drinks high in} \hlorange{sugar}\ctx{...} \\
\ctx{...\zh{小视频带您了解}}\hlorange{\zh{血脂}}\ctx{\zh{异常}...} \\
\ctx{...blood lipid (chole}\hlorange{sterol}\ctx{)...} \\
\ctx{...el 13\% de gr}\hlorange{asa}\ctx{, adem\'{a}s...} \\
\ctx{...the lowest} \hlorange{carbs}\ctx{?...}

\end{minipage}%
\hfill%
\begin{minipage}[t]{0.32\textwidth}
\raggedright

{\footnotesize\textbf{Transport Hubs}} \\[2pt]
\ctx{...\zh{从金钟地铁}}\hlred{\zh{站}}\ctx{\zh{游行至香港}...} \\
\ctx{...tram ride from Termini} \hlred{station}\ctx{...} \\
\ctx{...\zh{连接该校和红磡}}\hlred{\zh{火车站}}\ctx{...} \\
\ctx{...landing at Philadelphia} \hlred{Airport}\ctx{...} \\
\ctx{...the UNIL-Sorge} \hlred{station} \ctx{(8 stops)...}

\vspace{8pt}
{\footnotesize\textbf{Brand Names}} \\[2pt]
\ctx{...\zh{日本}}\hlviolet{\zh{迪士尼}}\ctx{\zh{商店推}...} \\
\ctx{...1968 Mercedes} \hlviolet{Benz}\ctx{...} \\
\ctx{...\zh{品牌：}}\hlviolet{\zh{联想}}\ctx{\zh{,惠普}...} \\
\ctx{...Zapatillas} \hlviolet{Adidas} \ctx{Mujer...} \\
\ctx{...the first} \hlviolet{McDonald}\ctx{'s...}

\end{minipage}

\caption{\textbf{Expert paths are monosemantic.} Each group shows tokens (highlighted) sharing an identical mid-layer expert path in \texttt{gpt-oss-20b}. Context is grayed; highlighted tokens are the routed unit. Paths cluster by function across languages and surface forms: \smash{\fcolorbox{black}{cyan!25}{\small overseas}}, \smash{\fcolorbox{black}{cyan!25}{\small\zh{海外}}}, and \smash{\fcolorbox{black}{cyan!25}{\small\zh{境外}}} share no lexical overlap but follow the same computation trace.
}
\label{fig:path-examples}
\end{figure*}

The results are striking (\Cref{fig:path-examples}). Tokens sharing a path form tightly coherent semantic groups that cut across languages and surface forms. One path clusters exhibition-related tokens (\zh{展}, \zh{展览}, \texttt{Exhibition}, \texttt{Showcase}); another groups \texttt{overseas}, \zh{海外}, and \zh{境外} despite no lexical overlap. Paths for nutritional substances group \texttt{cholesterol}, \texttt{sugar}, \zh{血脂}, and \texttt{grasa} across three languages; paths for brand names group \texttt{Adidas}, \zh{联想}, and \texttt{McDonald}. These clusters are representative of thousands of similarly coherent groups, and reflect \emph{what a token does in context}, not what it looks like---as previewed for the token \texttt{":"} in \Cref{fig:colon}, where identical surface form yields distinct paths for distinct functions.

\paragraph{The control subspace is the source.} Our decomposition predicts this: if paths compose control signals, then clustering in the control subspace should 
recover semantic structure, while clustering in content should not. We test this by independently clustering tokens in $h^{\text{vis}}_l$, $h^{\text{blind}}_l$, and $h_l$ at each layer ($k$-means, $k{=}32$ to match expert count), then grouping tokens that share the same cluster assignments across layers $8$--$16$.\footnote{Exact path matching applies only to $h^{\text{vis}}_l$; continuous clustering enables a fair comparison across subspaces.} We sample 100 cross-layer clusters from each subspace, filter those rated uninterpretable by automated scoring (${\sim}$5\%), and measure lexical diversity among the remainder.

All three subspaces produce interpretable clusters (${\geq}$95\% pass screening). The distinction is \emph{what organizes them}. $h^{\text{blind}}_l$ clusters are coherent but trivially so: they consist almost entirely of repeated instances of the same token (1.1 unique token IDs per cluster, out of 10 token samples per cluster). $h_l$ clusters show a similar pattern (1.4 unique IDs). $h^{\text{vis}}_l$ clusters, by contrast, group diverse tokens by shared function (4.3 unique IDs)---${4}{\times}$ the diversity of content clusters. Content preserves lexical identity; \textbf{control encodes semantic function}. The natural unit of interpretability in MoEs is not the expert---it is the trajectory.

\section{Discussion}
\label{sec:discussion}

MoEs separate computation into two orthogonal streams: a low-rank control signal that selects experts, and a content channel that carries the payload. Control rotates across layers while content accumulates, with each expert transforming today's content into tomorrow's control signal. This hand-off forces compositional specialization, producing expert paths that cluster tokens by semantic function across languages and surface forms.

\paragraph{Implications for neural control.} The control subspace is small, causally sufficient, and orthogonal to content---a natural target for 
intervention. Perturbing control could steer expert selection without corrupting content; modifying content could alter what is computed on while leaving routing 
unchanged. Whether this orthogonality survives into downstream behavior is an open question, but the decomposition provides the right coordinate system for asking it.

\paragraph{Broader perspective.} Dense models entangle ``what to compute'' and ``what to compute on'' in a single residual stream. MoEs, by introducing discrete routing, make this separation recoverable---and our results suggest the models themselves exploit it, composing simple control signals into rich semantic structure. We suspect the state--instruction duality is not unique to MoEs: dense models face the same organizational pressure, but without a discrete routing decision to make the separation identifiable. The decomposition tools may differ, but the question generalizes: understanding how representations \emph{control} computation, not just \emph{encode} information, may be essential to interpreting modern neural networks broadly.

% \begin{table}[H]
% \centering
% \small
% \caption{Top-1 expert prediction accuracy using top 2\% vs random 2\% of hidden dimensions (mid-layer).}
% \begin{tabular}{lcccccc}
% \toprule
% & Qwen3 & GPT-OSS & GLM-4.5 & OLMoE & Granite & DSv2 \\
% \midrule
% Top 2\% & \textbf{0.48} & \textbf{0.61} & \textbf{0.39} & \textbf{0.57} & \textbf{0.37} & \textbf{0.50} \\
% Random & 0.02 & 0.04 & 0.01 & 0.03 & 0.02 & 0.02 \\
% \bottomrule
% \end{tabular}
% \end{table}
% \begin{table}[H]
% \centering
% \small
% \caption{Top-1 expert prediction accuracy using top 2\% vs random 2\% of hidden dimensions (mid-layer).}
% \begin{tabular}{lcccccc}
% \toprule
% & Qwen3 & GPT-OSS & GLM-4.5 & OLMoE & Granite & DSv2 \\
% \midrule
% Top 2\% & \textbf{0.48} & \textbf{0.61} & \textbf{0.39} & \textbf{0.57} & \textbf{0.37} & \textbf{0.50} \\
% Random & 0.02 & 0.04 & 0.01 & 0.03 & 0.02 & 0.02 \\
% \bottomrule
% \end{tabular}
% \end{table}



\bibliography{iclr2026_conference}
\bibliographystyle{iclr2026_conference}

\appendix
\clearpage

\section{Routing Path Visualization}
\label{app:sugiyama}
We visualize expert paths as a layered directed acyclic graph, where nodes are experts at each layer and weighted edges represent token transitions between 
adjacent layers. The primary challenge is edge crossing: raw expert IDs are arbitrary, so even coherent routing structure appears as unintelligible spaghetti-plots when plotted with default ordering. We believe this visualization barrier is a key reason multi-layer routing paths have been underexplored despite their interpretive value.

We address this with a Sugiyama-style layered graph layout \citep{sugiyama1981}, a standard technique that reorders nodes at each layer to minimize edge crossings. We initialize expert positions by global usage frequency, then perform forward and backward barycenter sweeps: each expert's vertical position is updated to the weighted mean of its neighbors'positions, where weights reflect token flow. A single pass suffices for a stable layout.

Critically, we compute one shared layout from the \emph{aggregate} flow of all tokens, pooled across categories, and hold it fixed for all visualizations. In \Cref{fig:colon}, for example, the same expert ordering is used for all three panels. This ensures that visible differences in path coherence reflect genuine 
routing structure, not per-category layout optimization.

\section{Causally decomposing the residual stream}
\label{app:svd-decomp}
This appendix provides the full derivation of the decomposition introduced in \Cref{sec:svd-decomp}.

The residual stream in a MoE serves two distinct purposes. It is simultaneously a \textbf{control signal} that directs routing (\emph{what computation to perform}) and a \textbf{content payload} that is processed by the chosen experts (\emph{what data is operated on}). Here, we introduce a parameter-free decomposition based on SVD that precisely separates these two channels, splitting the pre-MLP residual stream $h_l$ into two orthogonal components:

\begin{align*}
  h_{l}
  \;=\;
  \underbrace{h^{\text{vis}}_{l}}_{\text{router–visible channel}}
  \;+\;
  \underbrace{h^{\text{blind}}_{l}}_{\text{router–blind channel}},
  \qquad
  \langle h^{\text{vis}}_{l},\,h^{\text{blind}}_{l}\rangle = 0.
\end{align*}

This allows us to \textbf{causally analyze what features drive expert specialization}. Due to the linearity of the routing gate, we can accomplish this without making modeling assumptions. We decompose $h_l$ into a router-visible channel that causally drives routing and a router-blind channel invisible to the router, as follows.

\paragraph{Methodology.}
A MoE router makes its routing decision at layer $l$ by computing the linear product of the routing matrix $R_l$ with the residual stream $h_l$:
\begin{align*}
    s_l = R_l^{\top}h_{l}.
\end{align*}
Crucially, this linearity means the router can only ``see'' information in $h_{l}$ that lies in the row space of its weight matrix $R_{l}\in\mathbb{R}^{N\times D}$.

We exploit this linearity via SVD to decompose the residual stream into two orthogonal components: what the router can see (router-visible) and what it cannot (router-blind). \Cref{fig:moe-diagram} illustrates how components flow through the MoE layer at inference time, when considered through the lens of this decomposition.

\begin{enumerate}
    \item \textbf{Computing the router's basis.}
    Let $R_{l}\in \mathbb{R}^{N\times D}$ be the routing matrix for layer $l$, where each row contains the weights for one of the $N$ experts. We compute its singular value decomposition:
    
    \begin{align*}
    R_{l} = U_{l}\,\Sigma_{l}\,V_{l}^{\top},
    \qquad
    \Sigma_{l}=\operatorname{diag}(\sigma_{1}\geq\cdots\geq\sigma_{r}>0),
    \end{align*}    
    
    where $r = \text{rank}(R_l) \leq \min(N, D)$. We retain only the $r$ right-singular vectors in $V_{l}\in\mathbb{R}^{D\times r}$ corresponding to non-zero singular values.

    \vspace{1.0em}
    \item \textbf{Orthogonal projectors.}
        The projector onto the router's row space is:
    
    \begin{align*}
    P_{l}=V_{l}V_{l}^{\top}\in\mathbb{R}^{D\times D},
    \qquad
    P_{l}^{2}=P_{l},\; P_{l}^{\top}=P_{l}.
    \end{align*}
    
    We decompose each residual stream as:
    
    \begin{align*}
    h^{\text{vis}}_{l}=P_{l}\,h_{l},
    \qquad
    h^{\text{blind}}_{l}=(I-P_{l})\,h_{l}.
    \end{align*}    
\end{enumerate}

By construction, $R_{l}h^{\text{blind}}_{l}=0$, meaning the router cannot access any information in $h^{\text{blind}}_{l}$. Thus, \emph{all routing decisions depend solely on $h^{\text{vis}}_{l}$}, while $h^{\text{blind}}_{l}$ is passed along with $h^{\text{vis}}_{l}$ to the routed experts for computation.

This decomposition mathematically guarantees that only $h^{\text{vis}}_{l}$ can causally influence expert selection; any relationship between $h^{\text{blind}}_{l}$ and routing outcomes is purely correlative, not causal.

\section{Models and Data}
\label{app:models}
To ensure our findings represent general principles of MoE design and not artifacts of a specific architecture, we analyze a diverse suite of high-performance models. Our selection spans different architectural families, attention mechanisms, parameter counts, and sparsity configurations. We test \texttt{Qwen3-30B-A3B} \citep{qwen32025}, \texttt{gpt-oss-20b} \citep{gptoss2025}, \texttt{GLM-4.5-Air} \citep{glm452025}, \texttt{OlMoE} \citep{olmoe2025}, \texttt{Granite-4-Tiny} \citep{granite2025}, and \texttt{Deepseek-v2-Lite} \citep{deepseekv22024}.

\paragraph{Models.}
\renewcommand{\arraystretch}{1.6}
\begin{table}[H]
\caption{Summary of models analyzed. $L$ = number of MoE layers (i.e., decoder layers excluding dense layers); $D$ = hidden dimension, $E$ = experts per MoE layer; $k$ = active experts per layer; $I$ = expert intermediate dimension (expressed as a fraction of $D$).}
\label{tab:model-summary}
\centering
\footnotesize
\setlength{\tabcolsep}{4.2pt}
\begin{tabular}{lccccccccc}
\toprule
Model & \makecell{Active/Total\\ Params} & $L$ & $D$ & $E$ & $k$ & $I$ & \makecell{Shared\\Experts} & \makecell{Notable\\characteristics} \\
\midrule
\textsc{Qwen3-30B-A3B} & 3B / 30B & 48 & 2048 & 128 & 8 & $\frac{3}{8}D$ & -- & Highly sparse activation \\
\textsc{GPT-OSS-20B} & 4B / 20B & 24 & 2880 & 32 & 4 & $D$ & -- & \makecell{Low-precision experts} \\
\textsc{GLM-4.5-Air} & 12B / 106B & 45 & 4096 & 128 & 8 & $\frac{11}{32}D$ & \cmark & Dense first layer \\
\textsc{OLMoE} & 1B / 7B & 16 & 2048 & 64  & 8 & $\frac{1}{2}D$  & -- & -- \\
\textsc{Granite-4-Tiny} & 1B / 7B & 40 & 1536 & 62  & 6 & $\frac{1}{3}D$ & -- & {Mamba2/Transformer hybrid}\\
\textsc{DeepSeek-v2-Lite} & 2B / 16B & 26 & 2048 & 64  & 6 & $\frac{11}{16}D$ & \cmark & \makecell{Multihead latent attention}\\
\bottomrule
\end{tabular}
\end{table}

All models are loaded with default bfloat16 weights, excluding \textsc{GPT-OSS-20B} (loaded with its recommended MXFP4 expert precision) and \textsc{GLM-4.5-Air} (loaded in quantized FP8). For each, we use the first supported attention implementation of FlashAttention-2 if supported by the model, and standard SDPA otherwise (except \textsc{GPT-OSS-20B}, which we load with recommended FlashAttention3).

\paragraph{Dataset.} 
All results from \Cref{sec:computation} use a 2 million token dataset sampled at sequence level from a mix of the \textsc{C4} \citep{c4} and \textsc{HPLTv2} \citep{hplt} datasets. For each model, we run forward passes, collect representations $h_l$ and expert selections at each layer $l$. For path analysis results, we use a 10m token dataset sampled from the same sources.

\section{Routing is driven by a low-dimensional control subspace}
\label{app:low-rank-routing}
This appendix provides full methodology and results for the signal amplification finding summarized in \Cref{sec:computation}.

We show that routing is a surprisingly low-dimensional process. MoEs learn a convergent strategy of \textbf{signal amplification}: routers learn to primarily listen to a tiny ``control subspace'' of dimensions that already have the highest magnitude, effectively amplifying the strongest features in the residual stream. We quantify this across layers and models.

\paragraph{Setup.}
To test this amplification hypothesis, for each layer $l$ and dimension $d \in \{1, \dots, D\}$ we measure:
\begin{enumerate}
  \item \textbf{Residual stream magnitude}. $M^h_{l, d} = \mathbb{E}_{tokens} \left[|h_{l, d}|\right]$: the average magnitude of representations $h_l$ at dimension $d$, averaged across $\sim$500k tokens sampled from standard text datasets (see \Cref{app:models}).
  \item \textbf{Router weight magnitude}. $M^R_{l, d} = \frac{1}{N}\sum_{e=1}^N |R_{l, d, e}|$: the average magnitude of router weights $R_l$ at dimension $d$, averaged across $N$ experts.
\end{enumerate}

We then compute the layer-wise Pearson correlation $\rho_l = corr(M^h_{l, \bullet}, M^R_{l, \bullet})$. A strong positive correlation would suggest routers implement an amplification dynamic, with routing driven by a small subspace of dimensions that already dominate the representation.

\paragraph{Findings.}
Across all models and layers, we observe strong positive amplification, with $\rho_l \approx 0.60$ on average. This pattern holds consistently across early and late layers, and from smaller to larger models, indicating a convergent strategy across diverse MoE architectures.

\begin{figure}[t] % htbp = placement hints: here, top, bottom, page
    \centering
    \includegraphics[width=0.9\linewidth]{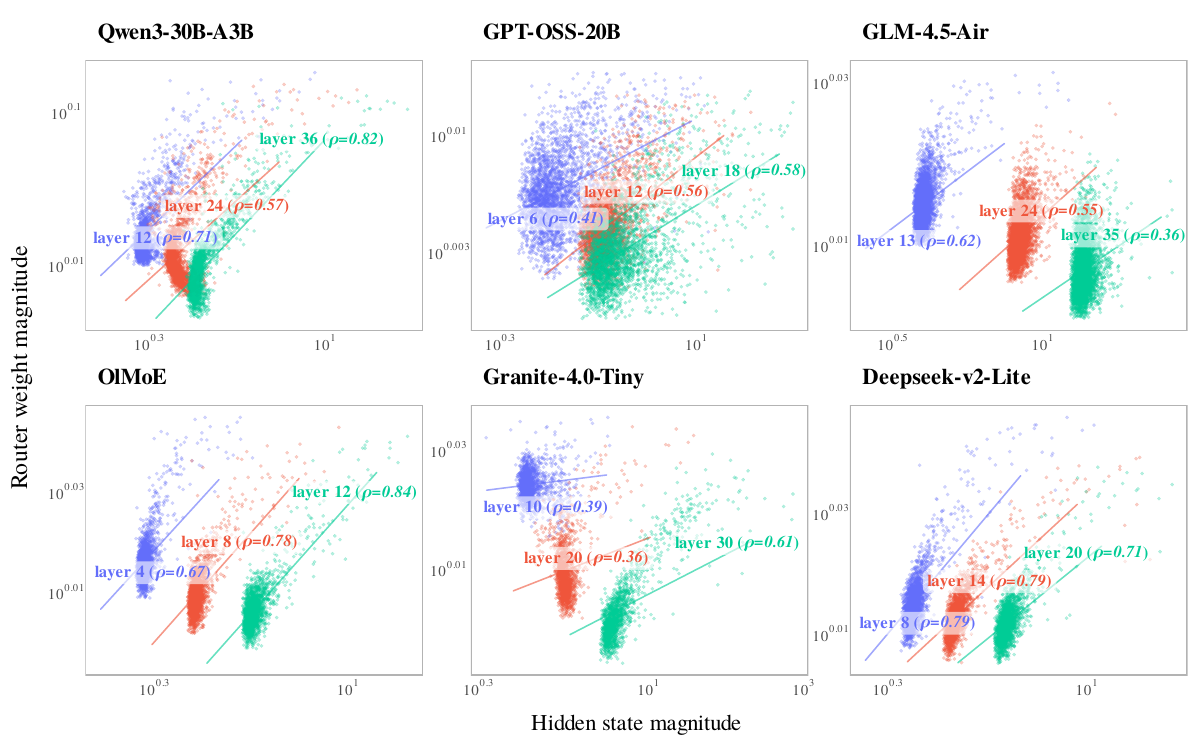}
  \caption{Per-dimension \textbf{hidden-state magnitude} $M^{h}_{l,d}$ (x-axis) versus \textbf{router-weight magnitude} $M^{R}_{l,d}$ (y-axis), both on log scales. One dot = one hidden dim~$d$ in layer~$l$. Colors represent layers. Strong positive correlations are observed, indicating that router weights are largest on dimensions where the representation already has the highest magnitude.}
   \label{fig:routing-dims} 
\end{figure}

\Cref{fig:routing-dims} visualizes this relationship for quartile layers (e.g., layers 12, 24, 36 in a 48-layer model), while \Cref{tab:routing-dims} provides more detailed values.

\renewcommand{\arraystretch}{1.1}
\begin{table}[t]
    \caption{Layer-by-layer $\rho_{l}$ values. Every fourth layer is listed for brevity; a -- indicates a MoE layer at that position does not exist in that model.}
    \label{tab:routing-dims}
    \centering
    {%
    \small
    \begin{tabular}{ccccccc}
        \toprule
         & \multicolumn{6}{c}{$\rho_{l}$} \\
        \cmidrule(lr){2-7}
        Layer & \textsc{Qwen3} & \textsc{GPT-OSS} & \textsc{GLM-4.5} & \textsc{OlMoE} & \textsc{Granite} & \textsc{DSv2-Lite} \\
        \midrule
         1  & +0.46 & +0.22 & --    & +0.70 & +0.22 & --    \\
         5  & +0.74 & +0.33 & +0.11 & +0.67 & +0.33 & +0.81 \\
         9  & +0.64 & +0.55 & +0.53 & +0.82 & +0.22 & +0.79 \\
        13  & +0.68 & +0.54 & +0.61 & --    & +0.57 & +0.81 \\
        17  & +0.73 & +0.56 & +0.61 & --    & +0.53 & +0.80 \\
        21  & +0.56 & +0.36 & +0.63 & --    & +0.46 & +0.72 \\
        25  & +0.54 & --    & +0.52 & --    & +0.79 & +0.63 \\
        29  & +0.82 & --    & +0.42 & --    & +0.83 & -- \\
        33  & +0.76 & --    & +0.39 & --    & +0.76 & -- \\
        37  & +0.87 & --    & +0.33 & --    & +0.69 & -- \\
        41  & +0.82 & --    & +0.38 & --    & --    & -- \\
        45  & +0.75 & --    & +0.42 & --    & --    & -- \\
        \bottomrule
    \end{tabular}
    }
\end{table}

This reveals that expert selection does not primarily rely on scanning the full $D$-dimensional space for subtle patterns, but instead operates within a low-dimensional, high-magnitude control subspace. The router's decision is thus driven less by a distributed signal network and more by the amplified signal of a few dominant features.

\paragraph{Causal validation.}
While these correlations are highly suggestive, they do not prove causation. Are these few dominant dimensions \emph{causally sufficient} to drive the routing decision? We test this by training simple probes to predict the top-1 expert ID using only a tiny fraction of the representation.

% We find that \textbf{a simple probe that sees only the largest 2\% of these dimensions can predict the router's expert choice with 45-60\% accuracy, despite choosing among 64-256 experts}.
%For a model with $D = 2048$, this means using just 40 dimensions to predict which of the 64-128 experts will be chosen.

For each layer $l$, we rank dimensions by their average magnitude $M^{h}_{l,d}$, then train two logistic regression probes:

\begin{enumerate}
  \item \textbf{High-mag probe}: uses only the highest-magnitude 2\% of dimensions in $h_l$.
  \item \textbf{Baseline probe}: uses a random 2\% of dimensions (results averaged across 10 random draws).
\end{enumerate}
% All regressions are trained on 80\% of the prior token samples, and evaluated on the held-out 20\%.

The results, shown in \Cref{tab:routing-acc-by-dim}, are unambiguous: a probe seeing only the top 2\% of dimensions can predict the router's choice with 30–80\% accuracy, while the random baseline remains near chance (1–2\%). For a model like \textsc{OLMoE} with $D=2048$, this means a probe using just 40 dimensions can predict which of the 64 experts will be top-1 with ~50\% accuracy.

This demonstrates that a small, high-magnitude subspace is not just correlated with routing decisions, but causally sufficient to drive them.

\renewcommand{\arraystretch}{1.1}
\begin{table}[H]
  \caption{Top-1 expert ID accuracy of a probe trained on the \textbf{top 2\% high-mag dimensions} (\textsc{High}) vs. a probe trained on a \textbf{random 2\% baseline} (\textsc{Base}). Every fourth layer shown for brevity; a -- indicates a MoE layer does not exist at that position in the model.}
  \label{tab:routing-acc-by-dim}
  \centering
  \small
  \setlength{\tabcolsep}{4.5pt}% tighter than the 6pt default

  \begin{tabular}{c cc cc cc cc cc cc}
    \toprule
          & \multicolumn{2}{c}{\textsc{Qwen3}}
          & \multicolumn{2}{c}{\textsc{GPT-OSS}}
          & \multicolumn{2}{c}{\textsc{GLM-4.5}}
          & \multicolumn{2}{c}{\textsc{OlMoE}}
          & \multicolumn{2}{c}{\textsc{Granite}}
          & \multicolumn{2}{c}{\textsc{DSv2-Lite}} \\
    \cmidrule(lr){2-3}\cmidrule(lr){4-5}\cmidrule(lr){6-7}\cmidrule(lr){8-9}\cmidrule(lr){10-11}\cmidrule(lr){12-13}
    Layer & \textsc{High} & \textsc{Base} & \textsc{High} & \textsc{Base} & \textsc{High} & \textsc{Base} & \textsc{High} & \textsc{Base} & \textsc{High} & \textsc{Base} & \textsc{High} & \textsc{Base} \\
    \midrule
     1  & \textbf{0.61} & 0.01 & \textbf{0.61} & 0.06 & -- & -- &\textbf{0.49} & 0.02 & \textbf{0.72} & 0.02 & -- & -- \\
     5  & \textbf{0.51} & 0.02 & \textbf{0.57} & 0.06 & \textbf{0.41} & 0.01 & \textbf{0.48} & 0.03 & \textbf{0.38} & 0.04 & \textbf{0.60} & 0.02 \\
     9  & \textbf{0.47} & 0.02 & \textbf{0.66} & 0.04 & \textbf{0.39} & 0.01 & \textbf{0.54} & 0.03 & \textbf{0.39} & 0.03 & \textbf{0.51} & 0.02 \\
    13  & \textbf{0.48} & 0.02 & \textbf{0.61} & 0.04 & \textbf{0.39} & 0.01 & \textbf{0.57} & 0.03 & \textbf{0.37} & 0.02 & \textbf{0.50} & 0.02 \\
    17  & \textbf{0.51} & 0.02 & \textbf{0.78} & 0.06 & \textbf{0.38} & 0.01 & -- & -- & \textbf{0.40} & 0.03 & \textbf{0.53} & 0.02 \\
    21  & \textbf{0.51} & 0.01 & \textbf{0.64} & 0.04 & \textbf{0.36} & 0.01 & -- & -- & \textbf{0.45} & 0.04 & \textbf{0.53} & 0.02 \\
    25  & \textbf{0.52} & 0.02 & -- & -- & \textbf{0.31} & 0.01 & -- & -- & \textbf{0.47} & 0.03 & \textbf{0.50} & 0.02 \\
    29  & \textbf{0.55} & 0.02 & -- & -- & \textbf{0.32} & 0.01 & -- & -- & \textbf{0.46} & 0.05 & -- & -- \\
    33  & \textbf{0.53} & 0.02 & -- & -- & \textbf{0.31} & 0.01 & -- & -- & \textbf{0.50} & 0.06 & -- & -- \\
    37  & \textbf{0.54} & 0.02 & -- & -- & \textbf{0.33} & 0.01 & -- & -- & \textbf{0.47} & 0.03 & -- & -- \\
    41  & \textbf{0.55} & 0.01 & -- & -- & \textbf{0.41} & 0.02 & -- & -- & -- & -- & -- & -- \\
    45  & \textbf{0.61} & 0.03 & -- & -- & \textbf{0.40} & 0.01 & -- & -- & -- & -- & -- & -- \\
    \bottomrule
  \end{tabular}
\end{table}

\begin{figure}[H] % htbp = placement hints: here, top, bottom, page
    \centering
    \includegraphics[width=.8\linewidth]{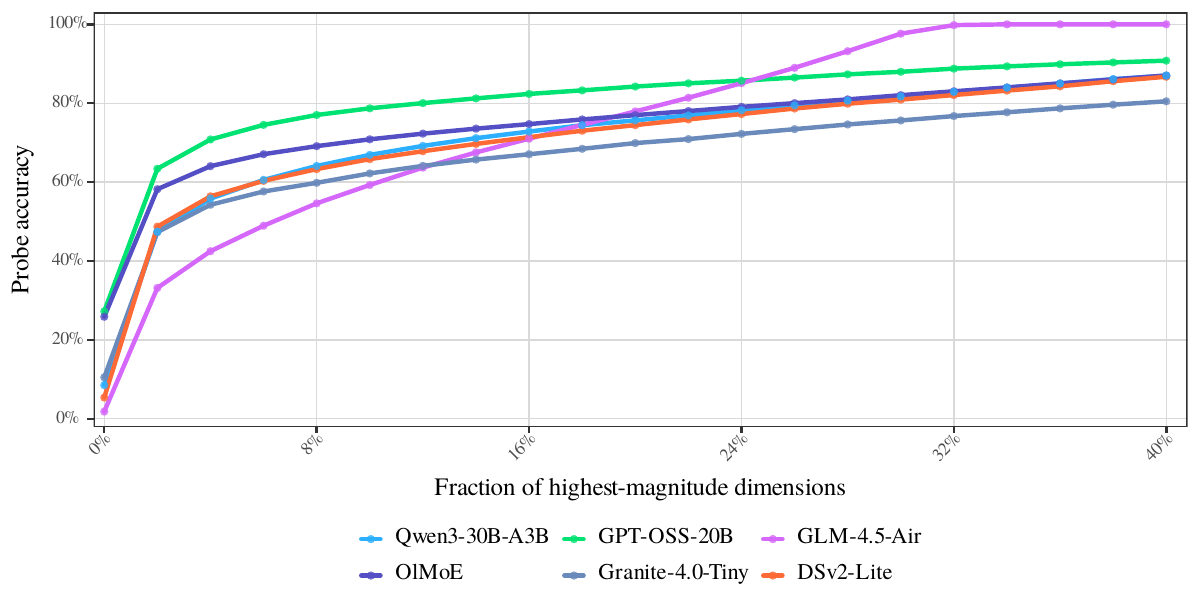}
  \caption{Probe accuracy versus fraction of highest-magnitude hidden dimensions used. Results shown for each model's midpoint layer. Accuracy saturates quickly, with the top 5\% of dimensions capturing the vast majority of routing information, demonstrating the low-rank nature of routing.}
   \label{fig:saturation-curve} 
\end{figure}

How many dimensions are truly necessary? As shown in \Cref{fig:saturation-curve}, probe accuracy saturates quickly, with rapidly diminishing returns after using the top 2-5\% of dimensions. This saturation profile is remarkably consistent across all tested models, despite significant variations in architecture.

\section{Feature rotation}
\label{app:feature_rotation}

\paragraph{Setup.} We use the same token samples from \Cref{app:low-rank-routing} and decompose their hidden states into the router-visible $h^{\text{vis}}_{l}$ and router-blind $h^{\text{blind}}_{l}$ channels. If routing signals are indeed transient, we expect the router-visible channel to be less stable across layers than the router-blind channel.

\begin{figure}[H]
    \centering
    \includegraphics[width=\linewidth]{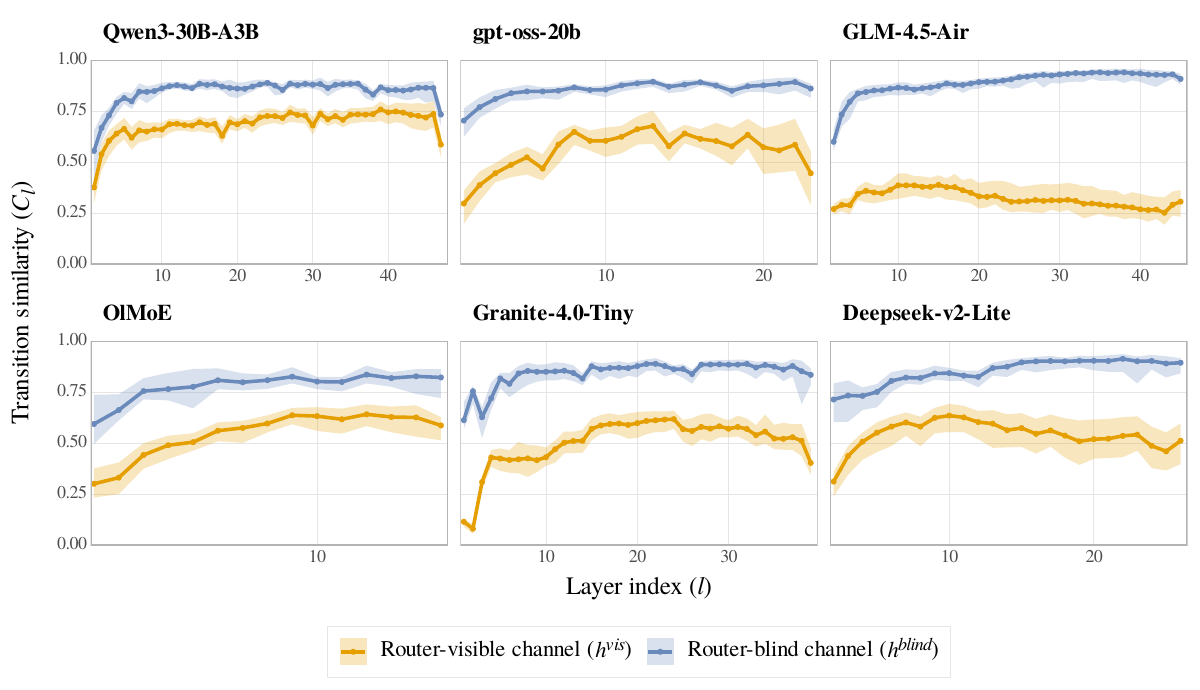}
  \caption{Average cross-layer cosine similarities $C_{l}^{\text{vis}}$ and $C_{l}^{\text{blind}}$ (shaded: 95\% bootstrap CIs).}
   \label{fig:channel-continuity} 
\end{figure}

For every adjacent pair of layers $(l,l\!+\!1)$ we measure the average cosine similarity within each channel:
\begin{align*}
  C^{\text{vis}}_{l}
  =\;
  \operatorname{cos}\bigl(h^{\text{vis}}_{l},\;h^{\text{vis}}_{l+1}\bigr),
  \qquad
  C^{\text{blind}}_{l}
  =\;
  \operatorname{cos}\bigl(h^{\text{blind}}_{l},\;h^{\text{blind}}_{l+1}\bigr).
\end{align*}

High similarity ($\approx 1$) indicates a stable, accumulating representation, while low similarity suggests a transient, rewritten signal.

\paragraph{Findings.}
\Cref{fig:channel-continuity} confirms a stark difference in stability. The router-blind channel ($h^{\text{blind}}$) is highly stable, with cross-layer similarity stabilizing at 0.75-0.90. This indicates that the information content not used for routing forms a continuous memory stream, accumulating and preserving features across layers.

In contrast, the router-visible channel ($h^{\text{vis}}$) is significantly less stable. This demonstrates that the specific features used to make a routing decision are transient and substantially change from one layer to the next.

\section{Channel probes}
\label{app:channel_probes}
We probe both channels for three surface features: language, token ID, and sequence position. We report normalized mutual information (MI\%) to enable comparison across features with different cardinalities.

\begin{figure}[H]
    \centering
    \includegraphics[width=.8\linewidth]{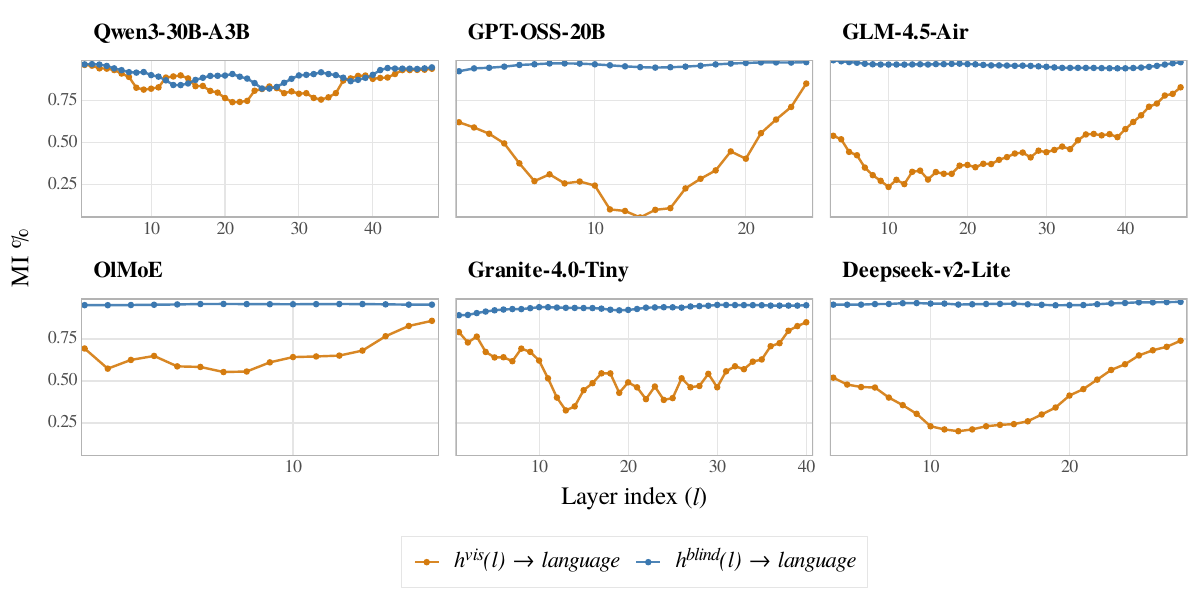}
    \caption{\textbf{Language probe.} Normalized MI\% between channel representations and language label. $h^{\text{blind}}_l$ (blue) encodes language consistently; $h^{\text{vis}}_l$ (orange) carries substantially less, especially at middle layers.}
\label{fig:language-probes}
\end{figure}

\begin{figure}[H]
    \centering
    \includegraphics[width=.8\linewidth]{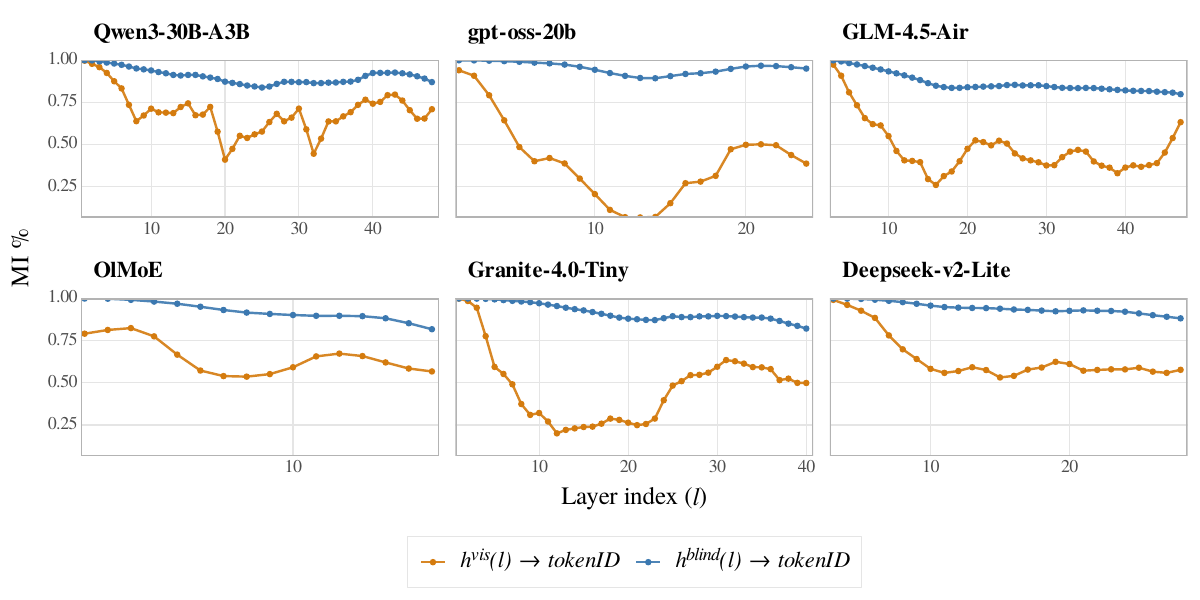}
    \caption{\textbf{Token ID probe results.} Normalized MI\% between channel representations and token ID (filtered for top 100 token IDs).}
    \label{fig:tid-probes}
\end{figure}

\begin{figure}[H]
    \centering
    \includegraphics[width=.8\linewidth]{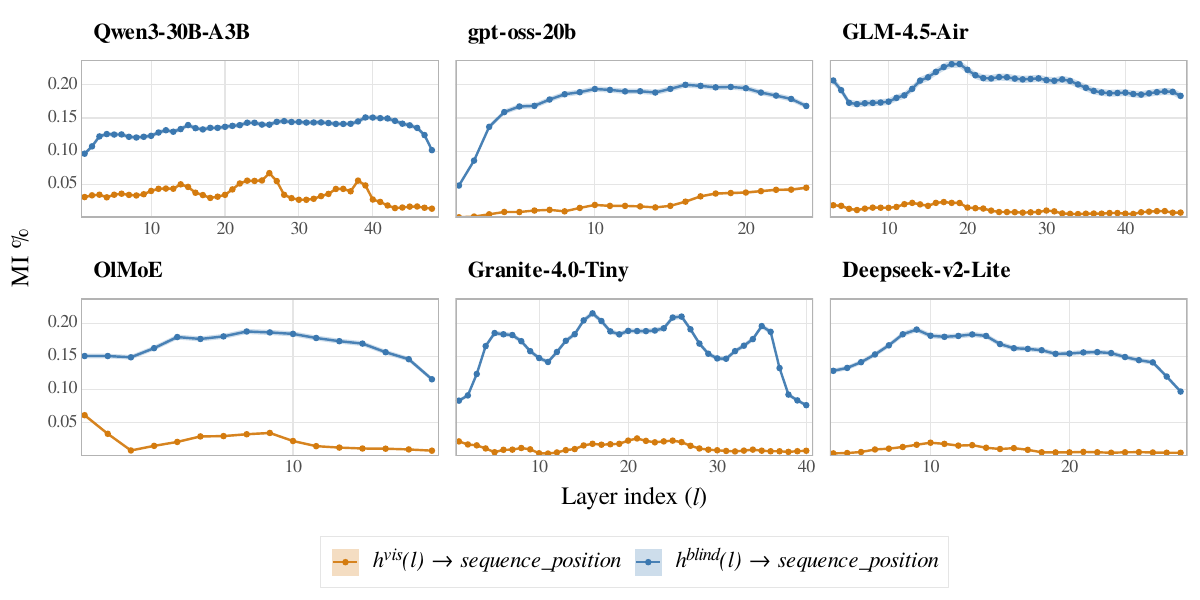}
    \caption{\textbf{Sequence position probe results.} Normalized MI\% between channel representations and token position ID.}
    \label{fig:position-probes}
\end{figure}

Across all six models and all three features (\Cref{fig:language-probes,fig:tid-probes,fig:position-probes}), the same pattern holds: surface features are encoded primarily in $h^{\text{blind}}_l$, with $h^{\text{vis}}_l$ carrying substantially less information. The disparity is largest at middle layers and narrows at the earliest and final layers, consistent with the interpretation discussed in \Cref{sec:computation}.

\section{Colon Token Dataset}
\label{app:colon-generation}
To generate balanced data for \Cref{fig:colon}, we iteratively prompt several auxiliary LLMs to produce text containing the token \texttt{":"} in each of three syntactic roles with high diversity, then randomly sample across auxiliary generators. Below are representative samples.

\paragraph{Type annotation.}~\\
\begin{tcolorbox}[colback=gray!5, colframe=gray!50, 
boxrule=0.5pt]
\small
def multiply\_values(a: int, b: int) -\textgreater{} int: return a * b
\end{tcolorbox}
\begin{tcolorbox}[colback=gray!5, colframe=gray!50, 
boxrule=0.5pt]
\small
function greetUser(name: string): string \{ return 
`Hello, \$\{name\}'; \}
\end{tcolorbox}

\paragraph{Introductory colon.}~\\
\begin{tcolorbox}[colback=gray!5, colframe=gray!50, 
boxrule=0.5pt]
\small
Please provide the following information: user ID, date 
of last login, and account status.
\end{tcolorbox}
\begin{tcolorbox}[colback=gray!5, colframe=gray!50, 
boxrule=0.5pt]
\small
The library's new guidelines are as follows: no food or 
beverages, no loud conversations, and no unapproved 
flyers.
\end{tcolorbox}

\paragraph{Time separator.}~\\
\begin{tcolorbox}[colback=gray!5, colframe=gray!50, 
boxrule=0.5pt]
\small
Meeting scheduled at 09:30 tomorrow. Please arrive by 
09:20 to set up.
\end{tcolorbox}
\begin{tcolorbox}[colback=gray!5, colframe=gray!50, 
boxrule=0.5pt]
\small
{[SERVER LOG]} 2025-07-12 14:03:52: User session 
started. Session validated.
\end{tcolorbox}

\section{LLM Usage}
We used an LLM for: (i) language editing and clarity and (ii) minimal code assistance. All technical content, experiments, and claims were implemented, checked, and verified by the authors. Any LLM-produced text/code was reviewed, edited, and validated; references and factual statements were manually verified. No LLM is an author; the authors take full responsibility for the content.\end{document}